\documentclass{article}

\usepackage[nonatbib, final]{nips_2016}

\usepackage[utf8]{inputenc} 
\usepackage[T1]{fontenc}    
\usepackage{url}            
\usepackage{booktabs}       
\usepackage{amsfonts}       
\usepackage{nicefrac}       
\usepackage{microtype}      
\usepackage{amsmath}
\usepackage{mathtools}
\usepackage{fancyvrb}
\usepackage{multirow}
\usepackage{color}
\usepackage{textcomp}

\DeclarePairedDelimiterX{\infdivx}[2]{(}{)}{%
  #1\;\delimsize\|\;#2%
}
\newcommand{\dkl}{D_\mathrm{KL}\infdivx}

\usepackage{listings}
\definecolor{lightgray}{rgb}{.9,.9,.9}
\definecolor{darkgray}{rgb}{.4,.4,.4}
\definecolor{purple}{rgb}{0.65, 0.12, 0.82}
\definecolor{orange}{rgb}{1,0.5,0}

\definecolor{Red}{RGB}{255,0,0}

\definecolor{Green}{RGB}{10,200,100}
\definecolor{Blue}{RGB}{10,100,200}

\lstdefinelanguage{JavaScript}{
  keywords={typeof, new, true, false, catch, function, return, null, catch, switch, var, if, in, while, do, else, case, break},
  keywordstyle=\color{blue}\bfseries,
  ndkeywords={class, export, boolean, throw, implements, import, this},
  ndkeywordstyle=\color{darkgray}\bfseries,
  identifierstyle=\color{black},
  sensitive=false,
  comment=[l]{//},
  morecomment=[s]{/*}{*/},
  commentstyle=\color{purple}\ttfamily,
  stringstyle=\color{red}\ttfamily,
  morestring=[b]',
  morestring=[b]"
}

\usepackage[ruled,vlined]{algorithm2e}

\DeclareMathOperator*{\argmax}{arg\,max}

\title{Practical optimal experiment design with probabilistic programs}

\author{
  Long Ouyang*, Michael Henry Tessler*, Daniel Ly*, Noah D. Goodman\\
  Department of Psychology\\
  Stanford University\\
  Stanford, CA 94305 \\
  \texttt{\{louyang, mtessler, lydaniel, ngoodman\}@stanford.edu}\\
}

\begin{document}

\maketitle

\begin{abstract}

Scientists often run experiments to distinguish competing theories.
This requires patience, rigor, and ingenuity---there is often a large space of possible experiments one could run.
But we need not comb this space by hand---if we represent our theories as formal models and explicitly declare the space of experiments, we can automate the search for good experiments, looking for those with high \emph{expected information gain}.
Here, we present a general and principled approach to experiment design based on probabilistic programming languages (PPLs).
PPLs offer a clean separation between declaring problems and solving them, which means that the scientist can automate experiment design by simply declaring her model and experiment spaces in the PPL without having to worry about the details of calculating information gain.
We demonstrate our system in two case studies drawn from cognitive psychology, where we use it to design optimal experiments in the domains of sequence prediction and categorization.
We find strong empirical validation that our automatically designed experiments were indeed optimal.
We conclude by discussing a number of interesting questions for future research.

\end{abstract}


\section{Introduction}
Designing scientific experiments to test competing theories is hard.
To distinguish theories, we would like to run experiments for which the theories make different predictions, but there are often many possible experiments one could run.
Formalizing theories as mathematical models can help.
Models make explicit hypotheses about observed data and thus make it easier (or in some cases, possible) to explore the implications of a set of theoretical ideas.
However, exploring models can be time-consuming and searching for experiments where the models sufficiently diverge is still largely driven by the scientist's intuition.
This intuition may be biased in a number of ways, such as towards experiments that show qualitative differences between models even when more informative quantitative differences may exist.


When the space of models and of experiments have been made explicit, it is possible to use optimal experiment design (OED) to \emph{automate} search; OED searches for experiments that maximally update our beliefs about a scientific question.
The information-theoretic foundations of OED are fairly straightforward \cite{Lindley1956}, but it has not enjoyed widespread use in practice.
Some OED systems are too narrow to be of general use and the more general systems require too much conceptual and implementational know-how to be widely adopted (e.g., users must supply their own objective function and derive a solution algorithm for it).
In order for OED to be both of general and practical use, the computation for experiment selection must be automatic.
This is only possible with a common formalism for specifying hypotheses and experiments.
\emph{Probabilistic programming languages} (PPLs) are such a formalism; they are high-level and universal languages for expressing probabilistic models.
In this work we describe a system in which the user expresses the competing hypotheses and possible experiments in a PPL; the optimal experiment is then computed with no further input from the user.

We first describe our framework in general terms and then apply it in two case studies from cognitive psychology.
Psychology is a good target domain for OED: hypotheses can often be expressed as mathematical models, rapid experiment iteration is possible and beneficial, and there is a large community ready to use (but not necessarily develop) sophisticated tools.
Psychological experiments also have certain challenging features: human participants give noisy responses, experimental results are sensitive to the size of one's sample, and computational models often do not make direct predictions about experimental data, instead requiring \emph{linking functions} to convert model output into predictions about data.
Our system naturally addresses these concerns.
In the first case study, we consider the problem of distinguishing three toy models of human sequence prediction.
In the second case study, we go beyond toy models and analyze a classic paper on human category learning that compared two models using an experiment designed by hand.
We find that OED discovers experiments that are several times more effective than the original in an information-theoretic sense.
Our work opens a number of rich areas for future development, which we explore in the discussion.

\section{Experiment design framework}
\label{s:bayes}
We begin with a concrete example before giving formal details.
Imagine that we are studying how people predict values for sequence data (e.g., flips of a possibly-trick coin).
We want to compare two models: $m_{\text{fair}}$, in which people believe the coin is unbiased, and $m_{\text{bias}}$, in which people believe the coin has a bias that is unknown (expressed as a uniform prior on the unit interval) but that can be learned from data.
We have a uniform prior on the models and we wish to update this belief distribution by conducting an experiment where we show people four flips of the same coin and ask them to predict what will happen on the next flip.
There are 16 possible experiments (all combinations of \lstinline{H} and \lstinline{T} for 4 flips) and $2^n$ possible outcomes---predictions of heads or tails for each of $n$ human participants.
Each model is a probability distribution on $\{0,1\}^n$ conditional on the experiment $x$---it describes a prediction about how people would respond after seeing some particular sequence of flips.
For convenience, we write our models in terms of a what a single person would do and assume that all people respond according to the same model, i.e., participant responses are i.i.d.\footnote{We use this simple linking function throughout this paper but our approach handles arbitrary linking functions (e.g., hierarchical models with subject-wise parameters).}

How informative would running the experiment \lstinline{HHTT} be?
$m_{\text{fair}}$ and $m_{\text{bias}}$ are identical here (heads and tails are equally likely)---if we ran our experiment with a single participant, neither experimental result would update our beliefs about the models, so this is a poor experiment.
By contrast, the experiment \lstinline{HHHH} would be much more informative.
Under $m_{\text{fair}}$, $p(\texttt{H}) = \frac{1}{2}$ but under $m_{\text{bias}}$, $p(\texttt{H}) = \frac{5}{6}$.
In this case, \emph{either} experimental response would be informative.
If the participant predicted heads, this would favor $m_{\text{bias}}$ and if she predicted tails, this would favor $m_{\text{fair}}$.
Thus, \lstinline{HHHH} would be a good experiment to run to disambiguate these models.
The goal of OED is to automate this reasoning.

We now formalize our framework.
We wish to compare a set of models $M$ in terms of how well they account for empirical phenomena.
A model $m$ is a conditional distribution $P_m(Y \mid X)$ representing the likelihood of empirical results $y$ for different possible experiments $x$.
We begin with a prior $P(M)$ and aim to conduct an experiment $x^*$ that maximally updates this distribution, providing as much information as possible.
That is, we wish to maximize $\dkl{ P(M \mid X = x^*) }{ P(M) }$.
\emph{A priori}, we do not know what the result of any particular experiment will be, so we must marginalize over the possible results $y$:
\begin{align}
  x^{*} &= \argmax_{x} {\mathbb E}_{p(y ; x)} \dkl{ P(M \mid X = x, Y = y) }{ P(M) }  \label{eq:oed}
\end{align}
where $p(y ; x)$ is the probability of observing result $y$ for experiment $x$.
If we have reason to believe that $M$ contains the true model of the data, then a suitable choice for $p(y ; x)$ is the predictive distribution implied by the models $p(y ; x) = {\mathbb E}_{p(m)} p_m(y \mid x)$.
If, however, we think $M$ may not contain the true model, then an uninformative prior $p(y ; x) \propto 1$ may be more appropriate.

\subsection{Writing models as probabilistic programs}

A key requirement for automating experiment design is to write hypotheses about the data as explicit models---in our case, probabilistic programs.
We use the probabilistic programming language WebPPL (\url{webppl.org}), a small but feature-rich probabilistic programming language embedded in Javascript \cite{dippl}.
WebPPL supplies a number of primitive distributions (e.g. \lstinline{Binomial}), which support generating samples and calculating the probability density of values from the domain.
For instance, we can sample from $\text{Binomial}(4, \frac{1}{2})$ using \lstinline|sample(Binomial({n: 4, p: 0.5}))| and we can determine the log-probability of the value 2 under this distribution using \lstinline|score(Binomial({n: 4, p: 0.5}), 2)|.
Often, we are interested in posterior inference. Let's say we are interested in the $\text{Binomial}(4, \frac{1}{2})$ distribution conditional on at least 2 successes. We write this as:
\begin{lstlisting}[mathescape, label={code:webppl}]
var g = function(){
	var x = sample(Binomial({n: 4, p: 0.5}))
	condition(x >= 2)
	return x
}
Infer(g) // returns the table p(2) = 6/11, p(3) = 4/11, p(4) = 1/11
\end{lstlisting}
\lstinline{g} is a function representing the conditional distribution.
Conceptually, it draws a sample \lstinline{x} from the prior, rejects values less than 2 using \lstinline{condition} (which enforces hard constraints\footnote{ An alternate form called \lstinline{factor} generalizes \lstinline{condition}, continuously weighting different program execution paths rather than simply accepting or rejecting them.}) and returns \lstinline{x}.
However, \lstinline{g} is not directly runnable.
To reify the conditional distribution, we must perform marginal inference on this model using \lstinline{Infer(g, options)}, which yields a distribution object (a probability table).
This provides a useful separation---we distinguish \emph{what} we wish to compute from \emph{how} we try to compute it.
In the above snippet, and throughout, we omit the \lstinline{options} object, which describes to \lstinline{Infer} which inference algorithm to use. WebPPL currently provides several inference algorithms: MCMC (MH, HMC), SMC, enumeration for discrete models, and variational inference.

\subsection{Writing OED as a probabilistic program}

Surprisingly, after expressing the spaces of models, experiments, and responses as probabilistic programs, it is straightforward to express OED as a probabilistic program as well (see Listing \ref{code:oed-pp}).
Equation~\ref{eq:oed} translates to around 20 lines of WebPPL code, expressing that OED is an inference problem.
A rich language like WebPPL is particularly well-suited here, as we lean heavily on the ability to perform \emph{nested} inference.
Also, writing OED as a probabilistic program gives us access to algorithms that are more sophisticated than previous research has considered (e.g., mixtures of enumeration and HMC for experiment spaces that have continuous and discrete subspaces).
Finally, note that we implement search for the optimal experiment using inference.
This is not essential---we could also replace the outermost \lstinline|Infer()| call with an optimization procedure (e.g., \lstinline|Search()|).

\begin{lstlisting}[mathescape, label={code:oed-pp}, caption = {OED implementation. For clarity, we have omitted some book-keeping details.}]
var OED = function(mSample, xSample, ySample) {
  var mPrior = Infer(mSample)             // store prior on models
  Infer(function() {                      // search over x
    var x = xSample()
    var KLDistrib = Infer(function() {    // compute KL for each y
      var y = ySample()                   // ${\color{purple} p(y ; x)}$
      var mPosterior = Infer(function() { // $\color{purple} P(M \mid Y = y)$
        var m = mSample()
        factor(score(m(x), y))
        return m
      })
      return KL(mPosterior, mPrior)      // ${\color{purple} \dkl{ P(M \mid Y = y) }{ P(M) } }$
    })
    var EIG = expectation(KLDistrib)     // ${\color{purple} {\mathbb E}_{p(y ; x)} \dkl{ P(M \mid Y = y) }{ P(M) } }$
    factor(Math.log(EIG / maxEIG))  // optional (search by inference)
    return {x: x, EIG: EIG}
  })
}
\end{lstlisting}
Our OED code is available as a WebPPL package at \url{https://censored}. 
We next illustrate our system by applying it to distinguish psychological theories of sequence prediction.

\section{Case study 1: Sequence prediction}
\label{s:tutorial}

Human judgments about sequences are surprisingly systematic and nonuniform across equally likely outcomes -- for example, we might strongly believe the next coin flip in the sequence \lstinline{HHTTHHTT} will be \lstinline{H}, whereas we might be unsure for the sequence \lstinline{THHTHTHT}.
There are many hypotheses one might have about what underlies human intuitions about such sequences \cite{goodfellow38:jep, falk81:pme, Griffiths2004_nips}.
Here, we consider three simple models of people's beliefs: (a) \emph{Fair coin}: people assume the coin is fair, (b) \emph{Bias coin}: people believe the coin has some unknown bias (i.e., the probability of a \lstinline{H} outcome) that they can learn from data, (c) \emph{Markov coin}: people believe the coin has some probability of transitioning between spans of \lstinline{H} and \lstinline{T} outcomes, also learnable from the data.
As in our earlier example, we consider an experimental setup where participants see four flips of the same coin and must predict the next flip.

\subsection{Formalization}

The model space $M$ is $\{m_{\text{fair}}, m_{\text{bias}}, m_{\text{markov}}\}$.
For now, we assume that the experiment will include data from a single participant, so the experiment space $X$ is the Cartesian product $\{1\} \times \{\texttt{H}, \texttt{T}\}^4$ representing the fixed sample size of 1 and sequence space.\footnote{Our notion of ``experiment'' is quite general, including traditional components like stimulus properties (e.g., coin sequence) as well as other components like dependent measure and sample size.}
Finally, $Y$ is the response set $\{\texttt{H}, \texttt{T}\}^1$.

In $m_{\text{fair}}$, we model participants as believing that the coin has an equal probability of coming up heads or tails:
\begin{lstlisting}[upquote=true]
var fairCoin = function(seq) {
  Infer(function(){ return flip(0.5) })
}
\end{lstlisting}
Here, \lstinline{flip(0.5)} is shorthand for \lstinline|sample(Bernoulli({p:0.5}))|.
Note the type signature of this model---it takes as input an experiment and returns a distribution on possible results of that experiment.

In $m_{\text{bias}}$, people assume the coin has some unknown bias, learn it from observations, and use it to predict the next coin flip:
\begin{lstlisting}[upquote=true]
var biasCoin = function(seq) {
  Infer(function(){
    var w = uniform(0,1), flipCoin = function(){ return flip(w) }
    var sampledSeq = repeat(seq.length, flipCoin)
    condition(arrayEquals(seq,sampledSeq))
    return flipCoin()
  })
}
\end{lstlisting}
\lstinline{biasCoin} first samples a coin weight \lstinline{w} from a uniform prior, uses it to sample a sequence of flips, and conditions on this matching the observed sequence \lstinline{seq}.
Thus, it learns the likely coin weights to have generated the observed sequence, and makes a prediction about the next flip.

Finally, $m_{\text{markov}}$ (see supplement for code) assumes that the flips are generated by a Markov process where the first coin flip is uniform and subsequent flips have some probability \lstinline{p} of transitioning away from the previous value.
\lstinline{p} is learned from the data and used to predict the next flip.

\subsection{Predictions of optimal experiment design}

Using an uninformative prior for $p(y; x)$, we ran OED for three different model comparison: fair--bias, bias--Markov, and fair--bias--Markov and planned to collect data from 20 participants (rather than 1).
\begin{figure}[t]
(a) \underline{\textsf{Example input for fair-bias comparison:}}
\begin{lstlisting}
var n = 20;
var fairGroup = groupify(fairCoin), biasGroup = groupify(biasCoin)
OED({mSample: function() { uniformDraw([fairGroup, biasGroup]) },
     xSample: function() {
    		 { n: n, seq: uniformDraw(["HHHH","HHHT",...,"TTTT"]) }
   	 },
     ySample: function() { return randomInteger(n + 1) } })
\end{lstlisting}
(b) \underline{\textsf{Output:}}\\
\includegraphics[width=\columnwidth]{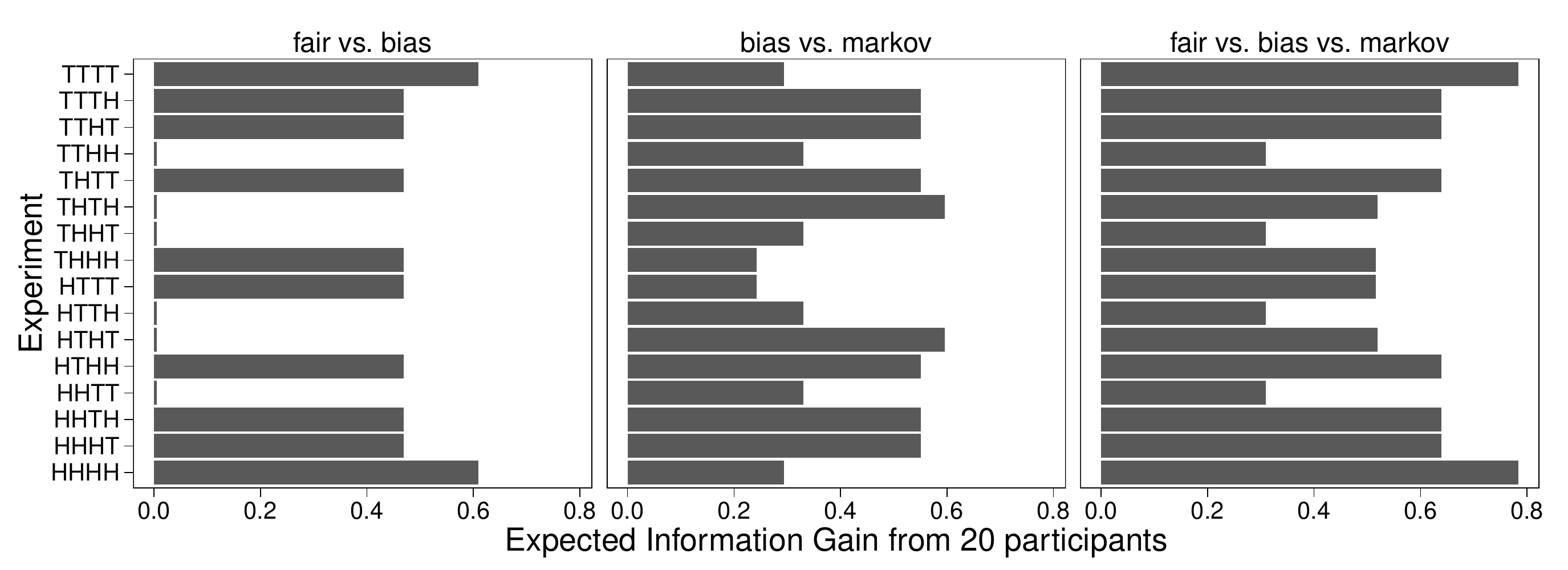}
\caption{(a) OED call and (b) results for sequence prediction model comparisons with 20 participants.}
\label{fig:run-coin}
\end{figure}

In the example call to \lstinline{OED} (Fig. \ref{fig:run-coin}a), we first lift each single-participant model into a model of group responses using an i.i.d. linking function \lstinline{groupify} (see supplement).
The experiment space \lstinline{xSample} includes a fixed number of participants and the unique sequences.
The response space \lstinline{ySample} is an uninformative prior over the number of \lstinline{H} responses.

Consider the fair--bias comparison (Fig.~\ref{fig:run-coin}b, left).
Observe that several experiments that have 0 information gain (e.g., \lstinline{HTHT}).
The models make exactly the same predictions in this case (albeit for different reasons), so the experiment has no distinguishing power.
The best experiments are \lstinline{HHHH} and \lstinline{TTTT}.
This is intuitive---the bias model would infer a strongly biased coin and make a strong prediction, while the fair coin model is unaffected by the observed sequence.

Moving to the bias--Markov comparison (Fig.~\ref{fig:run-coin}b, middle), the best and worst experiments actually reverse.
Now, \lstinline{HHHH} and \lstinline{TTTT} are the least informative (because, as before, the models make similar predictions here), whereas \lstinline{HTHT} and \lstinline{THTH} are the most informative.
This makes sense---the bias model learns a 0.5 probability of heads and so assigns equal probability of heads and tails to the next flip, whereas the Markov model learns that the transition probability is high and assigns high probability to the opposite of whatever outcome was observed last (\lstinline{T} for \lstinline{THTH} and \lstinline{H} for \lstinline{HTHT}).

\begin{figure}[h]
  \centering
  \begin{tabular}{l l}
    (a) & (b)\\
    \includegraphics[width=0.6\columnwidth]{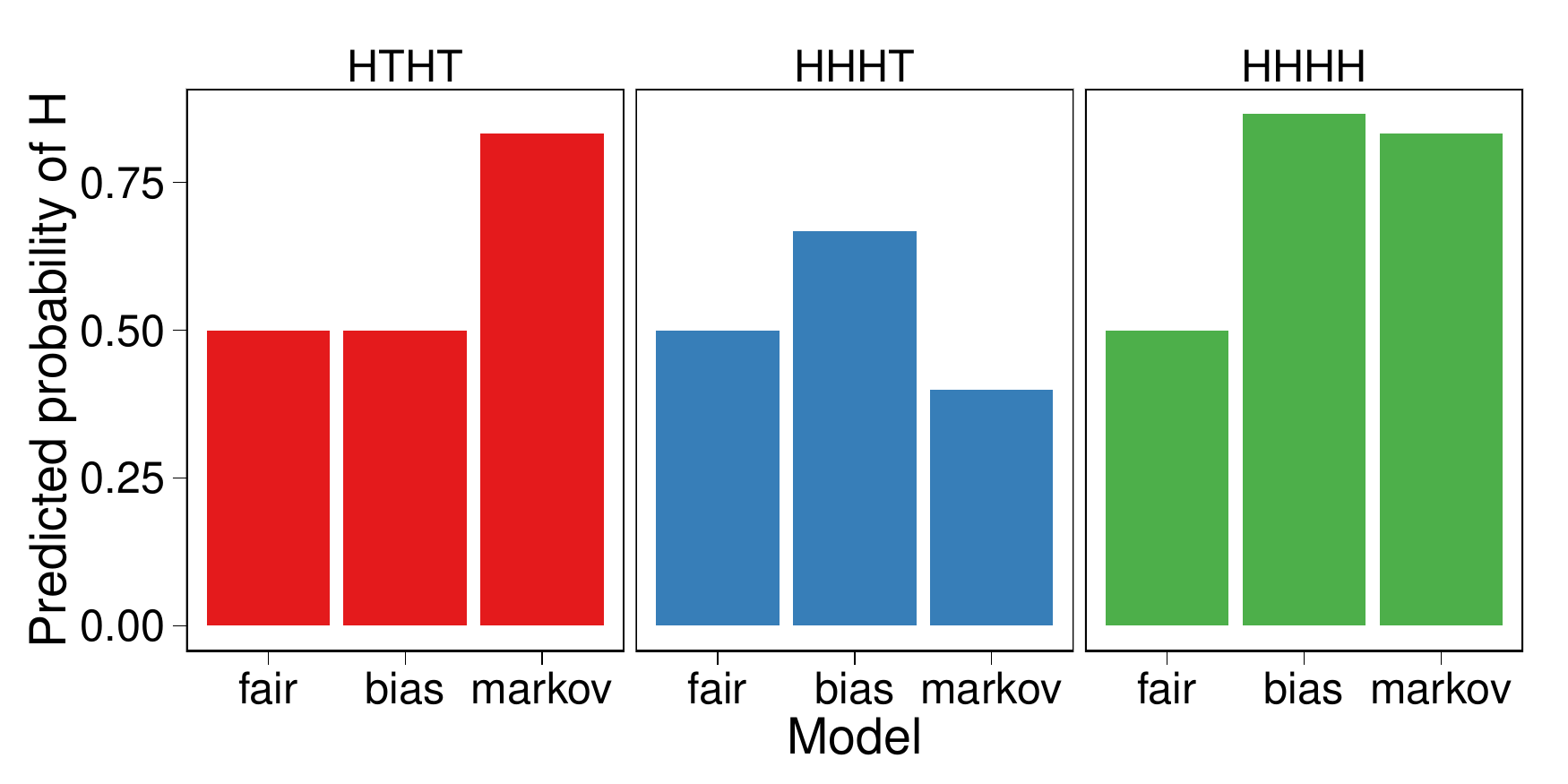} &
                                                                        \includegraphics[width=0.4\columnwidth]{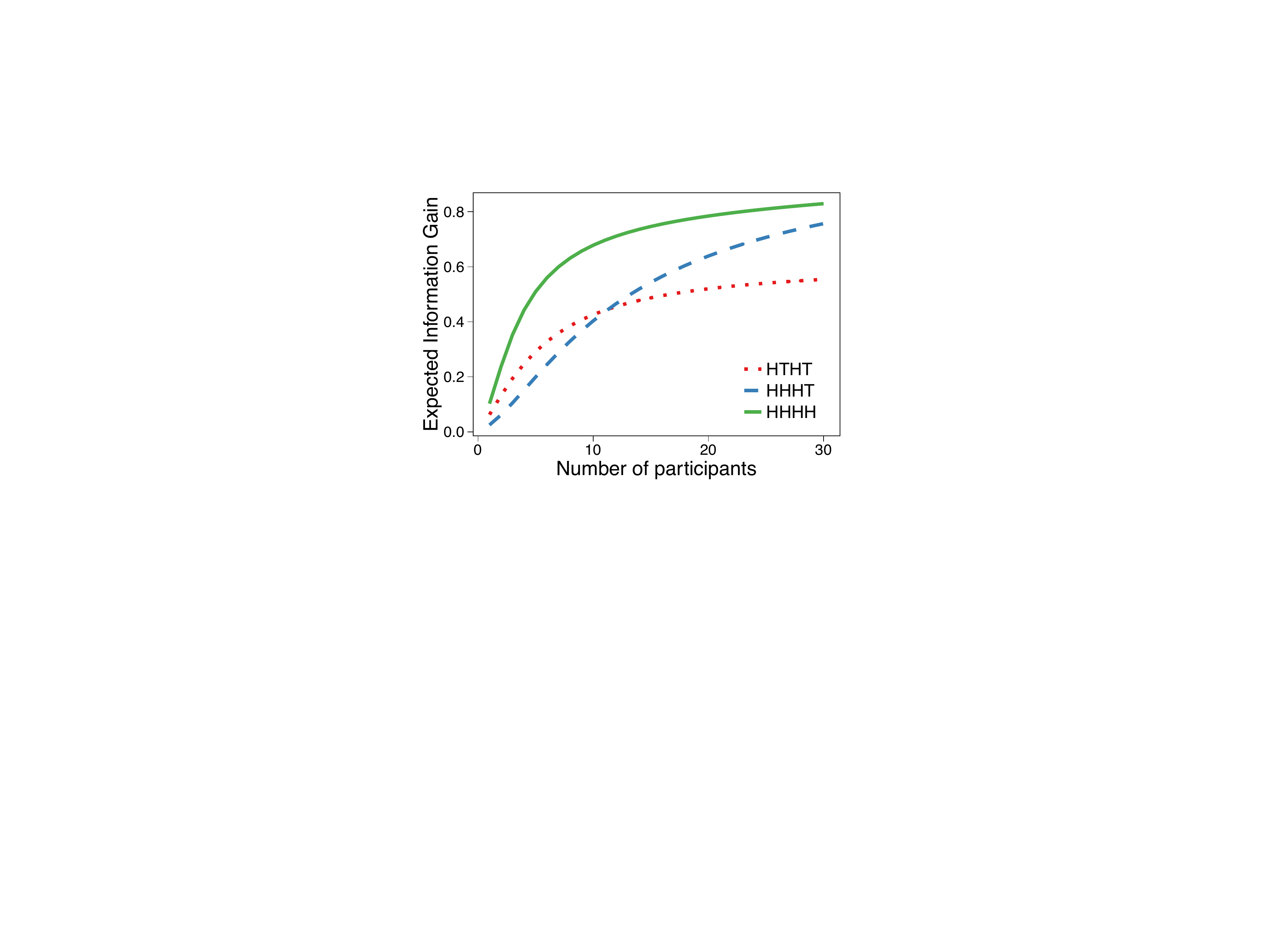} \\
  \end{tabular}
  \caption{(a) Model predictions for top three experiments (HHHH, HHHT, HTHT) in the full comparison (b) Expected information gain for these experiments versus sample size.}
  \label{fig:coin_preds}
\end{figure}

In the full bair--bias--Markov comparison (Fig.~\ref{fig:run-coin}b, right), the worst experiments (e.g., \lstinline{TTHH}) are again cases where all models make similar predictions.
The best experiments are \lstinline{TTTT} and \lstinline{HHHH}, a result that is non-obvious because we are comparing three models rather than two.
The best experiment \lstinline{HHHH} is very good at separating the fair model from the other two models, while still predicting a difference between bias and Markov (Fig.~\ref{fig:coin_preds}a, right).
The second best experiment, \lstinline{HHHT}, better distinguishes the bias model from the Markov model as it predicts a qualitative difference (Fig.~\ref{fig:coin_preds}a, middle), but this comes at the cost of less expected information gain overall.
An automated design tool is especially useful in these settings, where human intuition would likely favor the qualitative over the quantitative difference.

Finally, expected information gain of an experiment varies as a function of sample size (Fig.~\ref{fig:coin_preds}b).
This function is non-linear and, crucially, the rank ordering of experiments can change.
For the the full model comparison, the experiments \lstinline{HTHT} and \lstinline{HHHT} switch places after 12 participants.
This is particularly relevant when three models are being compared, as small quantitative differences between two models may amplify as the sample size grows.
In our example here, the optimal experiment with 1 participant is the same as with 30 participants.

\subsection{Empirical validation}
We validated our system by collecting human judgements for all 16 experiments and comparing expected information gain with the actual information gain from the empirical results.
We randomly assigned 351 participants to an experiment (all of the 16 experiments were completed by $\geq$20 unique participants).
Participants pressed a key to sequentially reveal the sequence of 4 flips and then predicted the next coin flip (either heads or tails).

\begin{figure}[h]
  \centering
  \includegraphics[width=0.7\columnwidth]{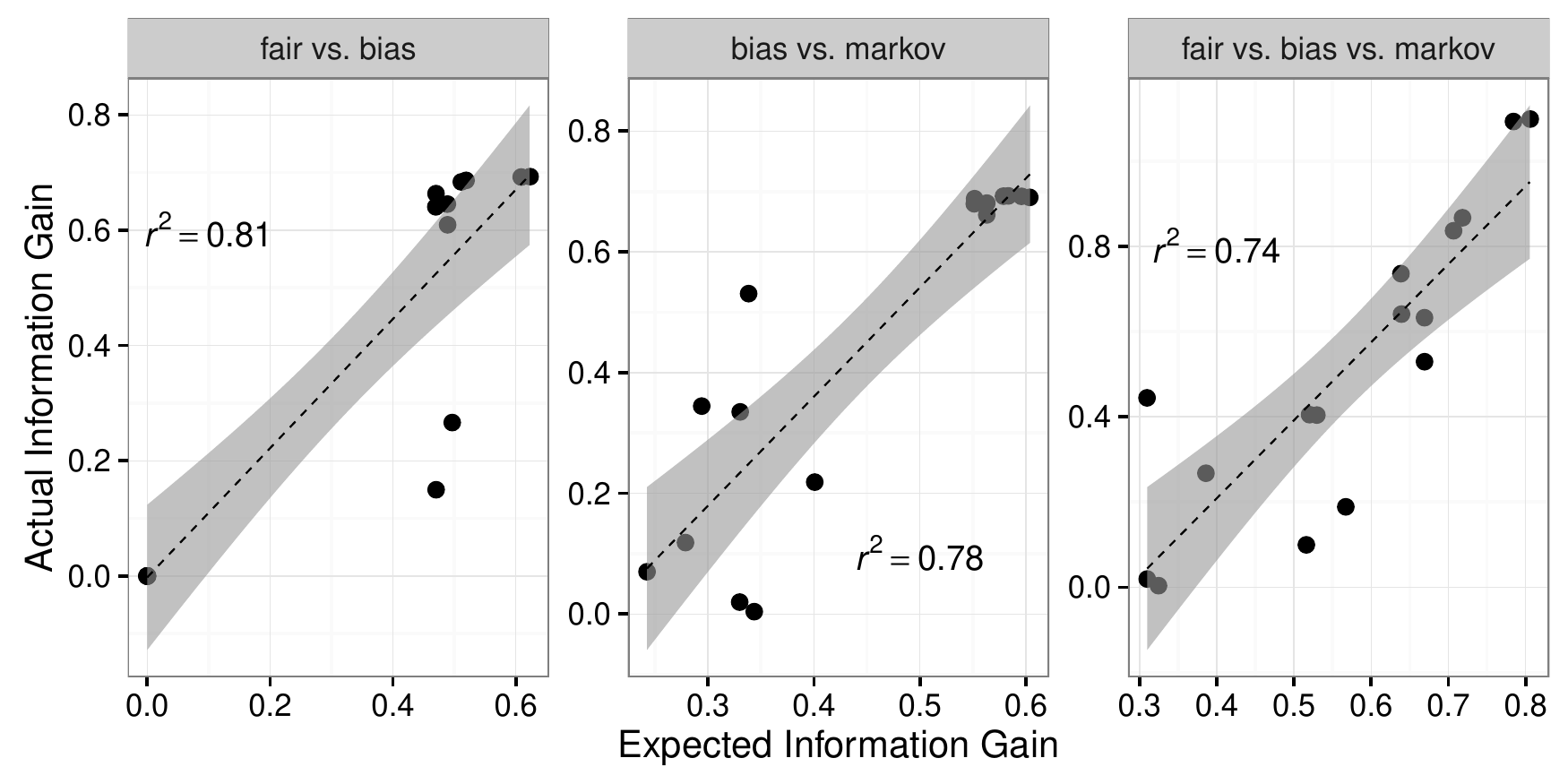}
  \caption{Actual vs. Expected Information Gain for each experiment}
  \label{fig:aig_vs_eig}
\end{figure}

For each experiment $x$ and result $y$, we computed the expected information gain from running our empirical sample of participants\footnote{N's are uneven due to randomization. We use the empirical N's for EIG in comparisons to AIG.} and compared this to the actual information gain, $\dkl{P(M \mid Y = y, X = x)}{P(M)}$, for the three model comparison scenarios.
Figure \ref{fig:aig_vs_eig} shows that expected information gain is a reliable predictor of the empirical value of an experiment (minimum $r$ = 0.857). This indicates that the OED tool could be relied on to automatically choose good experiments for this case study.

\section{Case study 2: Category learning}

Here, we explore a more complex and realistic space of models and experiments.
In particular, we analyze a classic paper on the psychology of categorization by Medin and Schaffer \cite{medin78:pr} that aimed to distinguish two competing models of category learning -- the \emph{exemplar model} and the \emph{prototype model}.
Using intuition, Medin and Schaffer (MS) designed an experiment (often referred to as the ``5-4 experiment'') where the models made diverging predictions and found that the results supported the exemplar model.
Subsequently, many other authors followed their lead, replicating and using this experiment to test other competing models.
Here, we ask: how good was the MS 5-4 experiment?
Could they have run an experiment that would have distinguished the two models with less data?


\subsection{Models}

Both the exemplar and prototype models are classifiers that map inputs (objects represented as a vector of Boolean features) to a probability distribution on the categorization response (a label: A or B).
The exemplar model assumes people store information about every instance of the category they have observed; categorizing an object is thus a function of the object's similarity to all of the examples of category A versus the similarity to all of B's examples.
By contrast, the prototype model assumes that people store a measure of central tendency for each category---a prototype.
Categorization of an object is thus a function of its similarity to the A prototype versus its similarity to the B prototype.
For details and representation of these models in WebPPL, see the supplement.

\subsection{Experiments}

Participants first learn about the category structures in a training phase where they perform supervised learning of a subset of the objects and are then tested on this learning in a test phase.
During training, participants see a subset of the objects presented one at a time and must label each object.
Initially, they can only guess at the labels, but they receive feedback so that they can eventually learn the category assignments.
After reaching a learning criterion, they complete the test phase, where they label all the objects (training set and the held out test set) without feedback.

MS used visual stimuli that varied on 4 binary dimensions (color: \emph{red} vs. \emph{green}, shape: \emph{triangle} vs. \emph{circle}, size: \emph{small} vs. \emph{large}, and count: \emph{1} vs. \emph{2}).
For technical reasons, they considered only experiments that (1) have linearly separable decision boundaries, (2) contain 5 A's and 4 B's in the training set, and (3) have the modal A object \lstinline{1111} and the modal B object \lstinline{0000}.
There are, up to permutation, 933 experiments that satisfy these constraints.

\subsection{Predictions of optimal experimental design}

\begin{figure}[t]
\centering
\begin{tabular}{l l}
(a) & (b)\\
\includegraphics[width=2.5in]{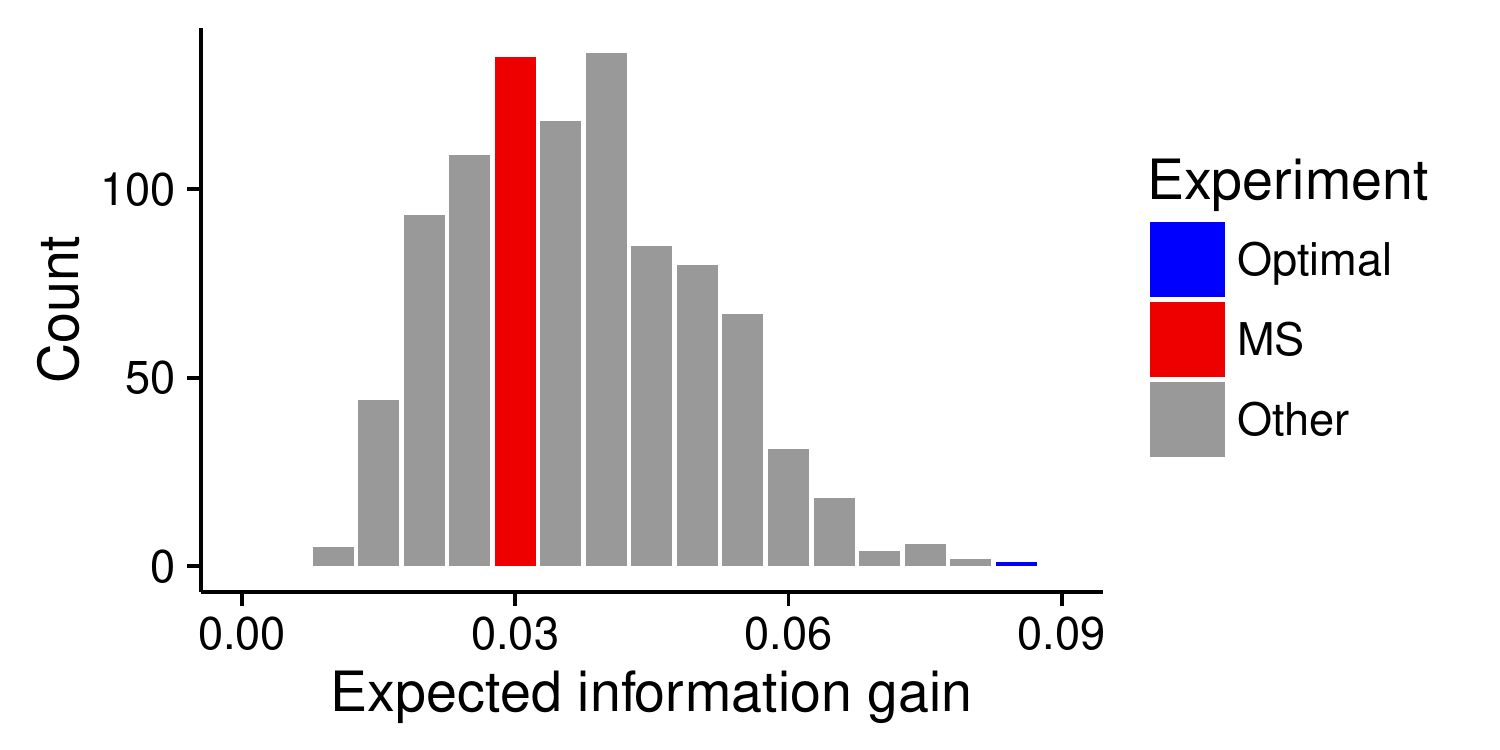} & \includegraphics[width=2.5in]{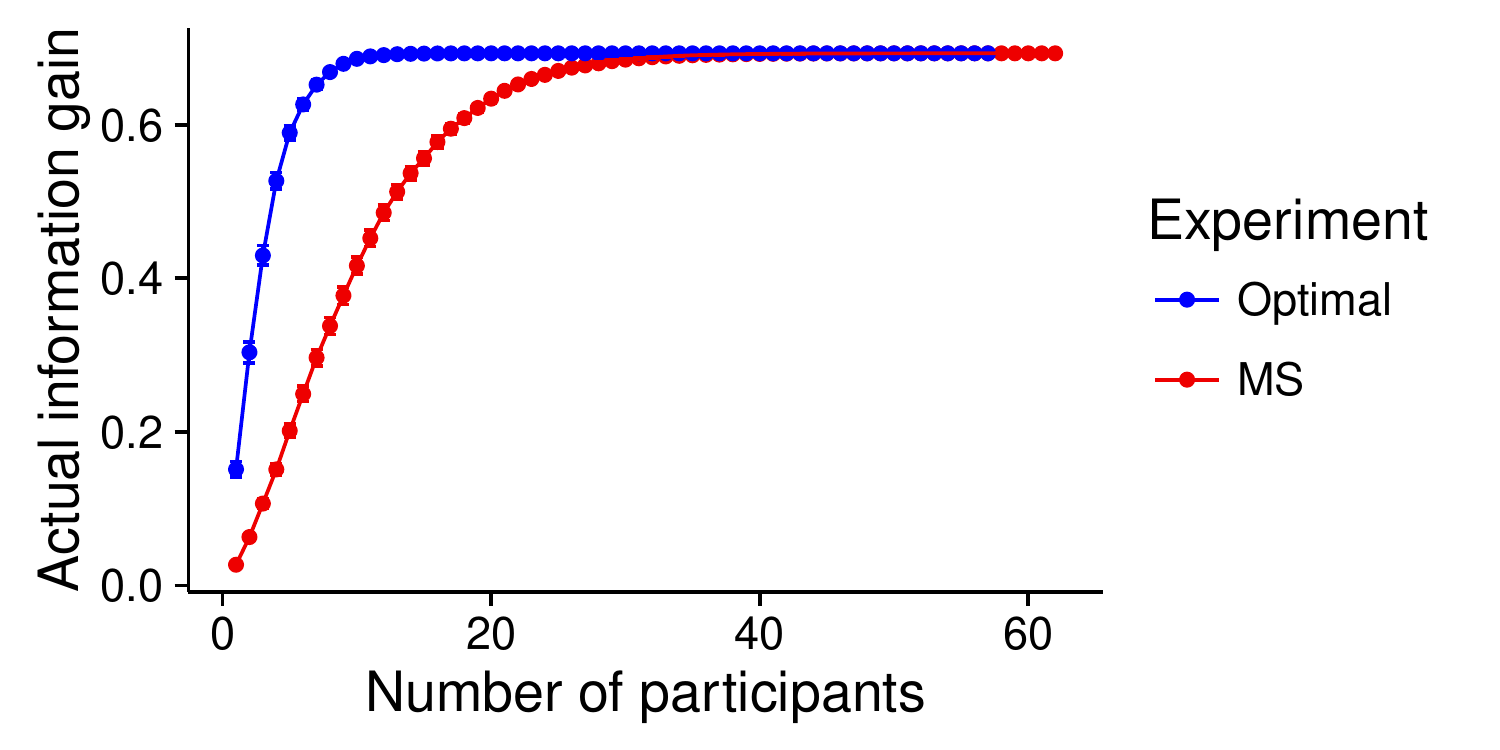}\\
\end{tabular}
\caption{(a) Distribution of expected information gain for all possible category learning experiments for a single participant. MS has low expected information gain. (b) Actual information gain versus number of experimental participants included in analysis (error bars are 95\% bootstrapped confidence intervals). MS requires three times as many participants to achieve maximum actual information gain.}
\label{fig:dist}
\end{figure}

Using the predictive prior for $p(y; x)$, we computed the expected information gain for all 933 experiments and found that the best experiment (for a single participant) had an expected information gain of 0.08 nats, whereas the MS 5-4 experiment had an expected information gain of only 0.03 nats.
Thus, the optimal experiment is expected to be 2.5 times more informative than the MS experiment.
Indeed, the MS experiment is near the bottom third of all experiments (Fig.~\ref{fig:dist}a).

Why is the MS experiment ineffective?
One reason is that Medin and Schaffer prioritized experiments that predict a qualitative categorization difference (i.e., when one model predicts that an object is an A while the other predicts it is a B).
The experiment they designed indeed predicts a qualitative difference for one object but this difference has a small magnitude and comes at the expense of little information gain from the remaining objects.
The optimal experiment is better able to quantitatively disambiguate the models by maximizing the information from all the objects simultaneously.

\subsection{Empirical validation}

To validate our expected information gain calculations, we ran the MS 5-4 and the optimal experiment with 60 participants each.
Figure~\ref{fig:dist}b shows that the optimal experiment we found for a single participant is indeed better than the MS experiment ($n$=1, blue greater than red).
For $n$=1, the mean actual information gain for the optimal experiment is 0.15, whereas it is 0.026 for the MS experiment.
This 5-fold difference in informativity is even greater than the 2.5-fold difference  predicted by expected information gain.
In addition, by incrementally introducing more data, we observe that both experiments achieve maximal actual information gain but the optimal experiment takes only 10 participants to asymptote to this maximum whereas the MS experiment takes around 30.
Thus, the optimal experiment provides the same amount of information for a third of the experimental cost.


\section{Related work}

The basic intuition behind OED---to find experiments that maximize some expected measure of informativeness---has been independently discovered in a number of fields, including physics \cite{vanDenBerg2003}, chemistry \cite{Huan2010}, biology \cite{Vanlier2012, Liepe2013}, psychology \cite{Myung2009}, statistics \cite{Lindley1956}, and machine learning \cite{Golovin2010}.

These papers, however, often implement OED for relatively limited cases, specializing to particular model classes and committing to a single inference technique.
For example, in systems biology, Liepe et al. \cite{Liepe2013} devised a method for finding experiments that optimize information gain for parameters of biomolecular models (ODEs with Gaussian noise).
Their information measure (Shannon entropy) is similar to ours, but they focus on a narrow family of models and commit to a bespoke inference technique (an ABC scheme based on SMC).
In psychology, Myung \& Pitt \cite{Myung2009} devised a general design optimization method but this method requires researchers to select their own utility function for the value of an experiment and implement inference on their own.
For example, they compared six memory retention models using Fisher Information Approximation as a utility function and performed inference using a custom annealed SMC algorithm.
Such ``bring-your-own'' requirements impose a significant burden on practitioners and are a real barrier to entry.

By contrast, we show that OED can be expressed as a generic, concise, and flexible function in a probabilistic programming language, which allows practitioners to rapidly explore different spaces of models, experiments, and inference algorithms.
Additionally, our work is the first to (1) demonstrate that expected information gain is a reliable predictor of actual information gain and to (2) characterize the cost benefits of OED.

\section{Conclusion}

Practitioners aim to design experiments that yield informative results.
Our approach partially automates experiment design, searching for experiments that maximally update beliefs about the model distribution.
With our approach, the scientist writes her hypotheses as probabilistic programs, sketches a space of possible experiments, and hands these to OED for experiment selection.
We stress that our work \emph{complements} practitioners; it does not replace them.
Our tool eliminates the need to manually comb large spaces for good experiments; we hope this will free scientists and engineers to work on the more interesting problems---devising empirical paradigms and building models.

Our approach suggests a number of interesting directions for future work.
We cast the OED problem as a problem of inference and this might suggest particular inference techniques.
For instance, if a particular response is quite unlikely (i.e., $p(y)$ is negligible), it may be acceptable to have a less precise estimate of information gain for that response.
Additionally, while our framework \emph{can} accommodate different costs of experiments using a prior distribution, the focus of our work is on finding \emph{informative} experiments.
It could be useful to integrate our system into multi-objective optimization systems for balancing multiple design considerations (e.g., informativeness, cost, ethics).
Finally, we have explored the trajectory of information gain as the number of \emph{i.i.d.} observations increases.
Observations need not independent, however.
Adaptive testing can be formulated as a problem of information gain of sequences of experiments, which produce dependent and non-identical responses.
In this paper, we showed case studies of our method in cognitive psychology but we believe that it is broadly useful, so we invite practitioners to test our method and system.

\bibliographystyle{ieeetr}
\bibliography{oed_nips_2016}

\end{document}